# An Inter-lingual Reference Approach For Multi-Lingual Ontology Matching

Haytham Al-Feel[1], Ralph Schafermeier[2] and Adrian Paschke[3]

[1] Information Systems Department , Fayoum University
Fayoum,  Egypt
*htf00@Fayoum.edu.eg*

[2] **AG Corporate Semantic Web, Freie Universität
Berlin, Germany**
*schaef@inf.fu-berlin.de*

[3] **AG Corporate Semantic Web, Freie Universität
Berlin, Germany**
*paschke@inf.fu-berlin.de*

**Abstract**
Ontologies are considered as the backbone of the Semantic Web. With the rising success of the Semantic Web, the number of participating communities from different countries is constantly increasing. The growing number of ontologies available in different natural languages leads to an interoperability problem. In this paper, we discuss several approaches for ontology matching; examine similarities and differences, identify weaknesses, and compare the existing automated approaches with the manual approaches for integrating multilingual ontologies. In addition to that, we propose a new architecture for a multilingual ontology matching service.  As a case study we used an example of two multilingual enterprise ontologies – the university ontology of Freie Universität Berlin and the ontology for Fayoum University in Egypt.

***Keywords:*** *Semantic Web, Ontology, Ontology Matching, Ontology Mapping, Multilingual Enterprises Ontologies.*

## 1. Introduction

Resources on the Semantic Web are designed to be accessible in a language-independent way. This is accomplished by using Unified Resource Identifiers (URIs) which serve as an ID for a Web resource, rather than a human-readable name. By definition, the local part of a URI can, but does not have to be human-readable. However, in order to provide humans access to Semantic Web resources using natural language, for example for querying a dataset using the SPARQL query language, resources can be enriched with human-readable annotations. For this purpose, RDF Schema defines the rdfs:label annotation type which allows to provide a human-readable version of a resource's name. In order to designate the language, a language tag can be added to the label.

Due to the prevalence of the English language in international research in general and in Semantic Web related research in particular, the majority of currently published Semantic Web resources are labeled using the English language. However, this is expected to change since more and more local communities have started publishing datasets labeled in their own language [1]. Semantic Web technology has the potential to provide semantic interoperability among heterogeneous systems which need to exchange information. A growing number of ontologies nowadays is built, which support different natural languages in order to address different cultures and roots. They often describe similar domains, but use different vocabularies. This poses new challenges for the multi-lingual integration of ontologies as well as the design of multi-lingual ontologies in general.

This leads to a growing relevance for multilingual ontology matching and mapping approaches, which, with the specific focus on multi-lingual ontology design and (re-)use, is relatively new research issue[2][3]. Examples are e.g., multilingual ontologies such as the Food and Agriculture Organization (FAO)[4], or semantic enterprise vocabularies which are used in cross-country business process, business rules or even country-specific standards which need to be shared due to the globalization of the economy.

In this research we will study the multi-lingual interoperability problem in enterprise ontologies. We analyze different ontology matching tools to see how monolingual or language agnostic ontology matching tools actually perform when dealing with multilingual ontologies. Therefore, we identify similarities and differences between them in order to facilitate the integration, collaboration and cooperation between different enterprises globally. As a case study for this research we use two multilingual enterprise ontologies that should be used to support the collaboration with each other; the first one is the ontology of Freie Universität Berlin in Germany which is expressed using German Language and the second one for Fayoum University in Egypt which is expressed in Arabic Language.

This paper proposes a general solution method for matching two specific multilingual ontologies through the translation to a third broadly accepted natural language. In our case study we suggest English natural language as the reference ontology language because it is accepted globally by different countries and cultures [1] . We will discuss several approaches for ontological matching to discover the suitability of these approaches for the matching of multilingual ontologies. This research can be extended in the future to enrich ICT services including e-learning initiative and digital library services that are provided to Egyptian Scholars and academic institutes [5].

This paper is organized as follows: Section 2 highlights the previous and related works. Section 3 describes the case study ontologies used in this research, showing their structure and characteristics. Section 4 discusses the different existing matching tools and the criteria for selecting a suitable matchmaking tool. Section 5 compares the existing automated approaches with the manual approach used as reference matching approach. Section 6 proposes a new architectural design for a multilingual ontology matching service. Finally Section 7 concludes our work and outlines the future work.

## 2. RELATED WORKS

Ontologies are considered as the main pillars in the Semantic Web and there are many ontology languages available such as RDF[6], RDFS[7], OWL[8], OBO[9], and the Ontology Definition Meta-model by the Object Working Group (OMG ODM)[10]. In addition, there is a lot of tool support, e.g. ontology editors such as Protégé[11], Neon Toolkit[12], Swoop[13], TopBraid Composer[14], and OntoStudio[15]. On the other hand, the support for the matchmaking of multilingual ontologies is considered as an urgent need in those tools, because most matching approaches have been developed for monolingual ontologies only and hence do not support the direct exchange of multilingual ontologies. With our approach of converting multilingual ontologies first to a reference standard ontology in English as common natural language before they are compared and matched, we establish a common ontological model on which the existing matchmaking tools can work.

There are a lot of efforts done by Euzenat in ontology matchmaking [16] [17] [18] [19]. Jorge Gracia and Eduardo Mena developed the CIDER matcher [20]. Dennis Spohr and his colleagues described the differences between monolingual, multilingual, and cross lingual ontologies from their point of view[21]. Mustafa Abusalah and his colleagues built ontology for the travel domain in Arabic and English languages and mapped them to each other via using the MRD dictionary [22] [23]. Bo Fu and colleagues proposed the SOMMO framework as a step towards multilingual ontology mapping [24].

On the other hand, there is not much research about indirect multilingual matching of ontologies. Jung and colleagues propose an indirect alignment of ontologies [25]. Their main idea is to reuse previously (manually or semi-automatically) obtained alignments between two respective ontologies in two different languages and a third intermediate ontology defined using a third language. Combining the two sets of alignments to a third set, yielding the appropriate alignments between the two ontologies of original interest. Espinoza and his colleagues described an approach to localize ontologies [26], while Tijerino described a cross functional implementation of cross lingual ontologies [27].

## 3. A CASE STUDY ON DEVELOPMENT & TRANSLATION OF MULTI-LINGUAL ONTOLOGIES

In this section we will introduce a use case with two ontologies, which we take as a basis for our further examination. Both ontologies represent two different organizations, namely Freie Universität Berlin and Fayoum University. They were modeled separately by the authors of this paper as domain experts having a different cultural and language background (German and Arabic). We used the Protégé ontology creation and editing tool. The natural languages used to build these ontologies are Arabic and German and these ontologies are represented in OWL. Figures.1 and 2 appears in the appendix show these two ontologies.

In our study, we then translated the two language dependent ontologies in a manual translation step into two ontologies in English, which serves as our common reference language. Therefore, the different primitives in both ontologies including the concepts, properties, and axioms were translated. We translated the local resource names as well as RDF:Labels. We added labels where none were present. This allows us to apply the existing ontology matching tools, which typically have been developed for ontologies in English language, to our translated ontologies. When we translated the different ontologies we did not change the structure of these ontologies. Furthermore, we took care of polysemy and synonym words [28]. Whenever possible, we also maintained the expressiveness and the type of the different primitives in the ontologies (e.g. a concept is mapped to a translated concept, an object properties is mapped to a translated object property etc.). This allows us in our evaluation to analyze string matching in addition to structure and semantic matchmaking. Table 1 gives a comparison of the two ontologies in our study.

---
[1] Besides that, most of the available ontologies are developed in English, which will make the process of interoperability between different enterprises much faster due to the number of already existing ontologies in English, which can be reused (at least partially).

Table 1: Properties of the two ontologies under study

| Criteria | Freie Universität Berlin ontology | Fayoum University ontology |
|---|---|---|
| Ontology Language | OWL 2 | OWL 2 |
| Total number of concepts | 70 | 187 |
| Total number of properties | 59 | 95 |
| Total number of primitives axioms (size of the ontology) | 192 | 523 |
| Total number of axioms | 663 | 2011 |
| Size of the ontology according to [29] | medium (101<500 concepts) | Large (501<1000 concepts) |
| Structure of the ontology according to [30] | simple (taxonomical tree structure) | simple (taxonomical tree structure) |
| Reuse/import of exiting ontologies | yes | No |
| DL Expressivity | SHOIQ(D) | ALCIN(D) |
| OWL Profile | OWL 2 DL | OWL 2 DL |

Both ontologies were developed independently of each other. However, they share a set of common properties at different levels. First, both ontologies cover the same domain, which is the organizational structure of a university.

None of the ontologies were developed following a specific ontology engineering methodology. However, the development processes of both ontologies underwent a requirements elicitation phase according to [31] which yielded a set of competency questions in each case. We were able to identify the following five competency questions which both ontologies should answer:

1. Does person A work at the university modeled by the ontology?
2. For which organizational unit does person A work?
3. Given a person A, which person B is the supervisor of A?
4. Given a person A, who are the co-workers of A in the same unit?
5. Given a unit A, which unit B is the super-unit of B?
6. Answer all subunits of a unit A.

For a further comparison of characteristics shared by both ontologies, we refer to Table 1.

## 4. EVALUATION OF MATCHING TOOLS FOR MULTI-LINGUAL ONTOLOGY MATCHMAKING

### 4.1 Ontology Matching

In this section, ontology matching and mapping can be done manually or by using automated tools. But, when the size and complexity of the ontology increases manual matching is no longer feasible. In this paper we matched the ontology both manually and by using ontology matching tools in order to have a comparison between both. In this section we will evaluate the capabilities of existing matchmaking tools for the match-making in the context of our multi-lingual ontology (translation) approach. We first describe the analysis criteria.

Ontology matching, as stated in [32] [21] [29], is defined as the process of finding relationships or correspondences between entities of two or more ontologies as an input and determine the output relationships between these entities. More formally, Euzenat defined ontology matching as "a function f that match a set of ontologies O1, … On, with an input alignment A, a set of parameters P, and a set of resources r, return an alignment A' between these ontologies A'=f{O1,….,On, A, P, r}" [19]

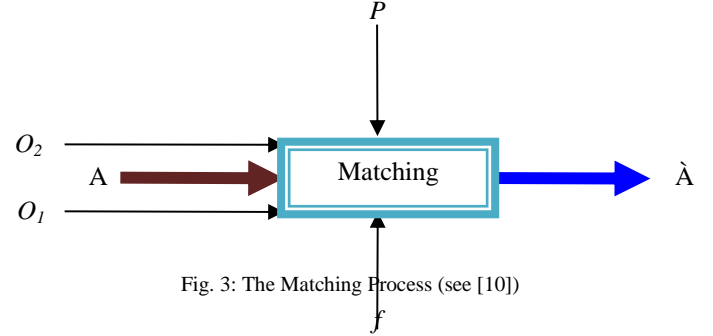

Fig. 3: The Matching Process (see [10])

While Ontology Alignment can be defined as "the correspondences between two or more ontologies as a result of the matching between these ontologies" [19]

There are many existing matching tools, such as COMA++ [33], Glue[34], FALCON[35][36], Prompt[37][38], and QOM[39] – many of them are just early proof-of-concept implementations which lack higher levels of software engineering quality. For our evaluation we selected the matchmaking tools from the surveys in [29] [40]. The criteria which we applied can be categorized according to dimensions such as the ontology format, the natural language that represents the ontology itself, the ontology size which varies between small (up to 100 primitives), medium (101-500 primitives), large (501-1000 primitives) and extra-large (over 1000 primitives) [29]. In addition, we also looked at the ontology domain and relations.

For the tool analysis we took the case study scenario and the competency questions listed in section III. Accordingly, the ontology alignment should contact areas of responsibility and the appropriate contact persons; we identify the problem as a specialization of the ontology matching problem.

As a first step, we conducted an investigation of the capabilities of existing matching tools with respect to our requirements given by our scenario. Due to the vast amount of different ontology matching approaches and tools available, we decided to follow a defined selection process.

First, we classified existing ontology matching approaches according to a set of relevant criteria as in Table II. Since both ontologies studied in this paper are expressed using the Web Ontology Language OWL 2, we were confined to matching tools capable of dealing with OWL 2 ontologies. Furthermore, a number of matching tools were designed to deal with ontologies covering restricted domains, such as the bio-medical domain. In order to work with our university ontologies, we were limited to domain-agnostic matchers. The last two criteria reflect the simple practical implications that the tools we investigate be usable out-of-the-box and publicly available.

Applying these criteria leaded the following set of selected ontology matching tools:
Falcon
RiMOM
Anchor-Flood
AgreementMaker

Table 2: Classification of Existing Ontology Matching Approaches

| Matcher | Ontology representation language | Domain-specific | Stand-alone | Availability |
|---|---|---|---|---|
| **Falcon** [42] | OWL | no | Yes | Open Source |
| SAMBO [43] | RDFS/OWL | yes (biomedicine) | Yes | - |
| **RiMOM** [44] | OWL | no | Yes | Open Source* |
| **Anchor-Flood** [45] | RDFS/OWL | no | | Open Source |
| ASMOV [46] | OWL | yes (biomedicine) | | - |
| DSSim [47] | OWL/SKOS | no | no (part of AQUA question answering system [48]) | Not available |
| AgreementMaker [49] | XML/RDFS/OWL/N3 | no | | Needs registration (request pending) |

* Not able to get it to run.

## 4.2 Matching Results

Falcon was only able to match nine concepts, no individuals, and no object or data type properties. Two of the nine matches are considered false positives with respect to our reference alignment as shown in Figure 4.

Fig. 4: Results of the Ontology Alignment Using Falcon

Anchor-Flood detected 12 concept matchings, four of which were false positives. No matching at the instance level or at object or data type property level was detected.

RIMOM on the other hand detected 72 matchings, but with only 6 being correct and 66 being false positives (see Figure 5).

Fig. 5: Results of the Ontology Alignment Using RIMOM

Our explanation for these bad results in matching comes from the translation of ontologies. Although, they are covering roughly the same domain and are annotated with labels in English, the underlying original ontologies represent concrete institutions with similar but different hierarchical structures. Furthermore, many similar or even equal concepts are in both ontologies, the terminology used to name these concepts differs in their semantic meaning in many instances. For instance, in cases where a 1:1 equality mapping is not appropriate because two concepts from both ontologies are similar but not equal or in cases where one or more concepts from one ontology can be subsumed by a concept in the other ontology, it would be useful to achieve an m:n mapping, which none of the tools under study are capable of.

Another shortcoming in the analyzed ontology matching approaches is the inability of concept-to-instance matching. It is, for example, possible and common that the same role is modeled as a concept in one ontology while it is modeled as an instance in another ontology and as an object property in a third ontology. These and similar alternatives for modeling one and the same phenomenon is following different Ontology Design Patterns [50].

In the two example ontologies used in this paper, there exist many examples where things are modeled as classes in one ontology and as instance data in the other. The inability of the matching tools to identify this use of different ontology design patterns has a significant impact on the result of the matching process, which is that a high number of potential matches remain undetected.

For this reason, we claim that class-instance matching has the potential of improving matching results significantly in situations where two ontologies using different development patterns need to be matched.

## 5. AN ARCHITECTURE FOR A MULTILINGUAL ONTOLOGY MATCHING SERVICE

Our proposed architecture for a multilingual ontology matching service is depicted in Figure 6. It builds on top of a generic classical hybrid matching architecture, consisting of string based/linguistic, structural, and/or semantic matching services.

We extend this architecture by a translation service that translates the labels for concepts, individuals, and properties of one ontology into the language of the other, so as to allow string based and linguistic matching. In addition to the translation service, a word sense disambiguation service is needed in case homonymy, homography, or polysemy is encountered during the translation process.

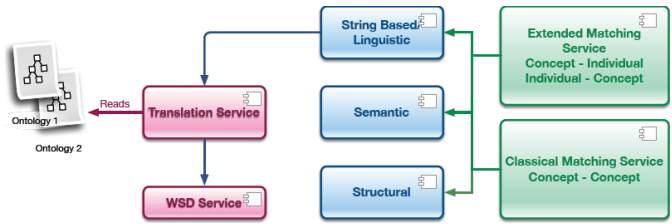

Fig. 6: Architecture for a Multilingual Ontology Matching Service

The second extension consists of an enhanced matching service that does not only match equal or similar primitives of the same kind (concepts, individuals, and properties) but also permits matching concepts to individuals and vice versa. As described in section 4 we have found evidence of different modeling patterns used in different ontologies for expressing the same thing. This poses a potential problem for the matching process and is not unique to but prevalent in ontologies modeled by different parties using different languages. As the experiment described in section 4 shows, matching results could be significantly improved by allowing for the matching of concepts to individuals and vice versa.

This architecture is one of several building blocks of our general architecture for the inter-lingual Semantic Web. In a next step, we will supplement our work by defining architecture for inter-lingual access to data and meaningful information on the semantic web.

## 6. CONCLUSION & FUTURE WORK

It can be foreseen that in the next years more and more ontologies will be multilingual and will be published on the Semantic Web. The problem of multi-lingual matching and integration has been only rarely studied in literature. Our study of this interoperability problem on the example of matching two multilingual enterprise ontologies and our evaluation of different ontology matching tools highlighted some weakness. There are several available matching tools but they do not support direct multi-lingual ontology matching, since they have been developed for the matching of monolingual ontologies. In our work we proposed the translation of multi-lingual ontologies into ontologies which make use of a common natural language such as English for their descriptions, so that the existing monolingual matcher can be applied for the matching of them. But even in case of perfectly translated ontologies into a common language model, we discovered several shortcomings in recent ontology matchers, such as the inability of finding correct matchings in case the ontologies have been modeled differently (e.g. different ontology structure, different concept semantics etc.). Derived from our case study and tool analysis we defined a general process for the multi-lingual transformation pre-processing and matchmaking and mapped it into a general component architecture which implements the process steps. Our future work will propose solutions for address certain shortcomings in the discussed tools such as the class-instance matching.

The authors of this paper participated in the development of the Arabic chapter of DBpedia and part of them participated in the German chapter. From this point of view we are concerned with the internationalization and the multilingual problem of the Semantic Web and DBpedia. In this paper we studied if it is better for DBpedia to use different ontologies in different natural languages in the same domain mapped to each other or if it is better to use only one ontology model that has different datasets in different languages mapped to the same ontology (e.g. by an additional language label for each concept and property). We proposed a reference architecture for the first scenario, but, based on the results of our research studies, for the purpose of a multi-lingual DBPedia we suggest adding more properties to cover number of attributes available in different languages which have no equivalent properties according to the cultural perspective.

**Appendix**

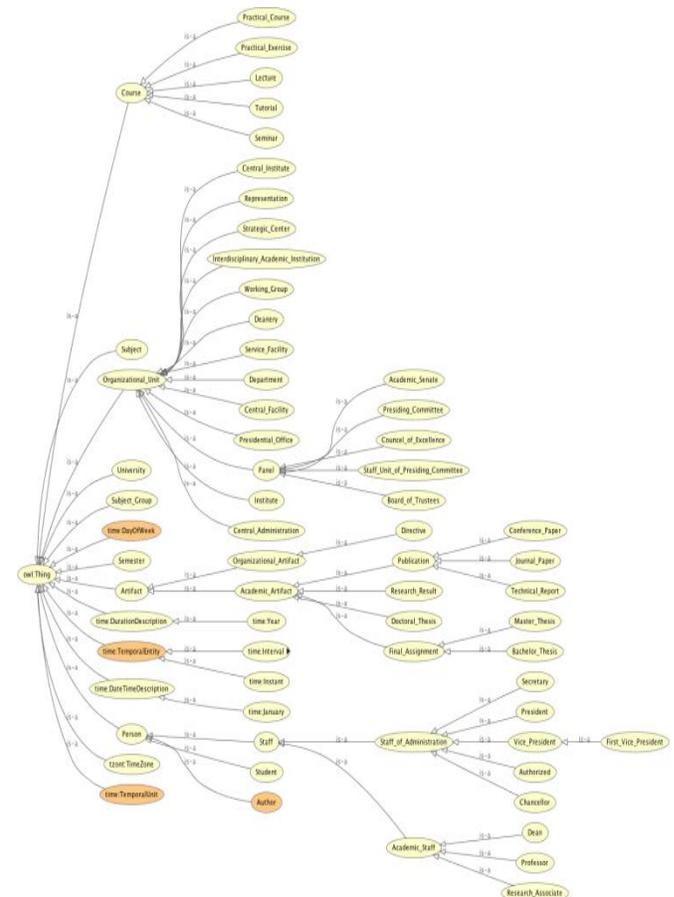

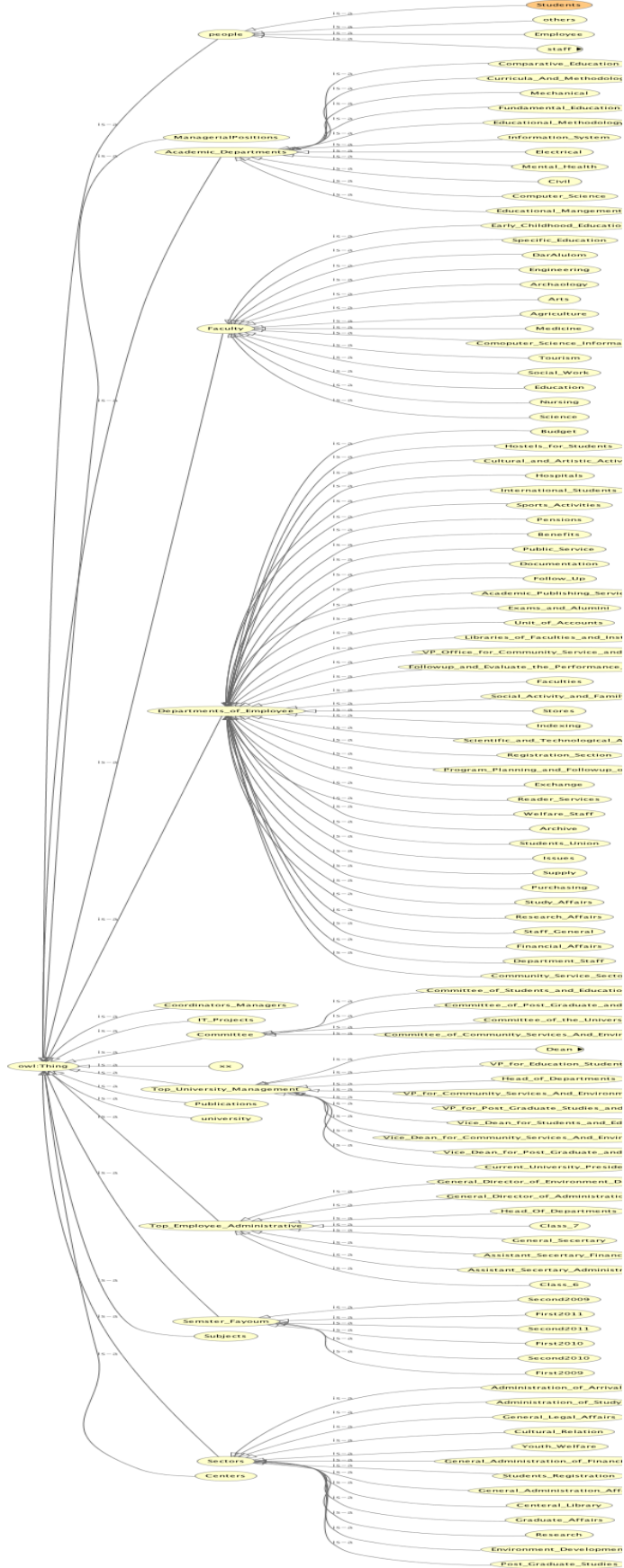

Fig. 1: Excerpt from the Freie Universität Ontology

Fig. 2: Excerpt from the Fayoum University ontology